\documentclass[10pt,twocolumn,letterpaper]{article}

\usepackage{cvpr}              
\usepackage{arydshln}
\usepackage{multirow}
\usepackage{adjustbox}

\definecolor{cvprblue}{rgb}{0.21,0.49,0.74}
\usepackage[pagebackref,breaklinks,colorlinks,allcolors=cvprblue]{hyperref}

\usepackage{pifont}
\newcommand{\cmark}{\ding{51}}%
\newcommand{\xmark}{\ding{55}}%

\title{\raisebox{-12pt}{\includegraphics[width=0.08\textwidth]{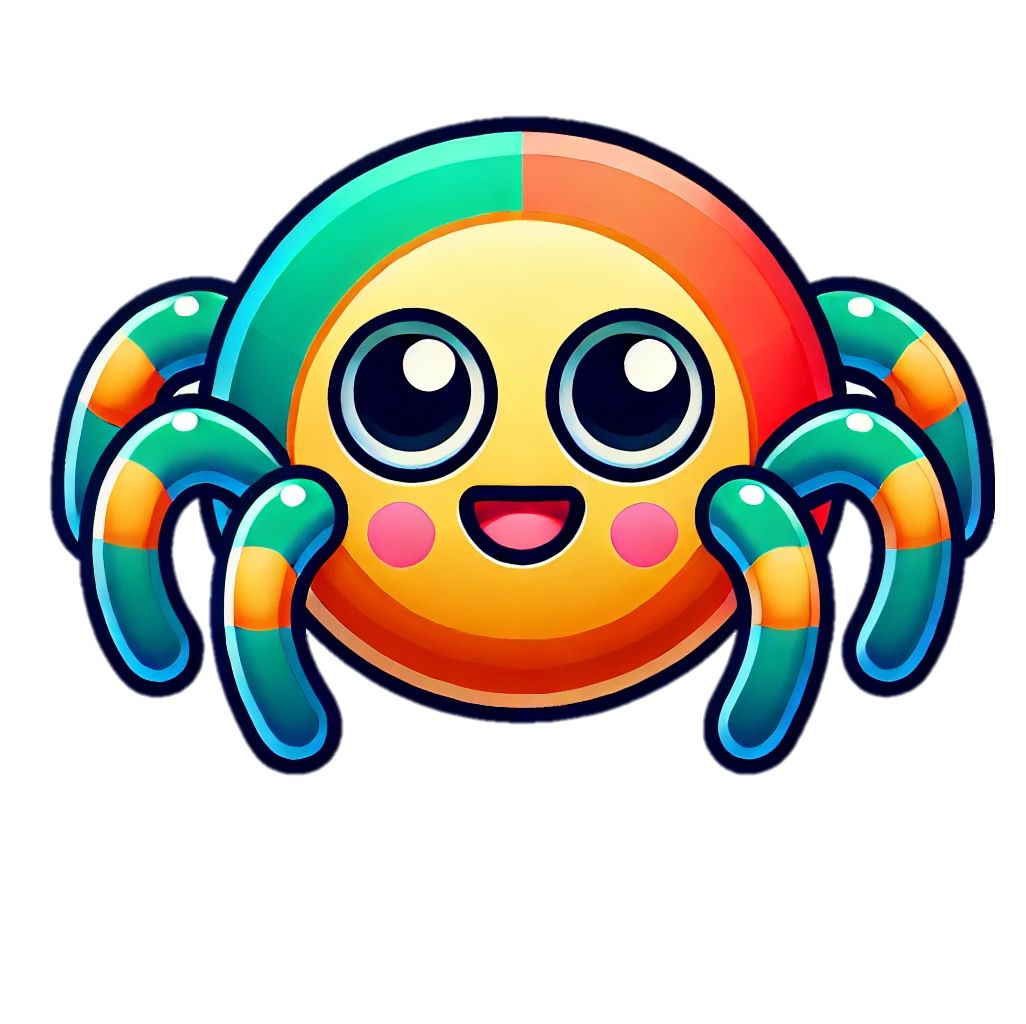}} Dispider: Enabling Video LLMs with Active Real-Time Interaction via \\ Disentangled Perception, Decision, and Reaction}

\author{    
    \textbf{Rui Qian}$^{1}$\thanks{Equal Contribution} ~
    \textbf{Shuangrui Ding}$^{1}$\footnotemark[1] ~
    \textbf{Xiaoyi Dong}$^{1,2}$ ~
    \textbf{Pan Zhang}$^2$ ~ \\
    \textbf{Yuhang Zang}$^2$ ~
    \textbf{Yuhang Cao}$^2$ ~
    \textbf{Dahua Lin}$^{1,2}$ ~
    \textbf{Jiaqi Wang}$^2$\\
    $^1$\ The Chinese University of Hong Kong\\
    $^2$\ Shanghai AI Laboratory\\ 
    {\tt\small \{qr021, ds023\}@ie.cuhk.edu.hk}
}

\begin{document}
\maketitle

\begin{abstract}
Active Real-time interaction with video LLMs introduces a new paradigm for human-computer interaction, where the model not only understands user intent but also responds while continuously processing streaming video on the fly.
Unlike offline video LLMs, which analyze the entire video before answering questions, active real-time interaction requires three capabilities: 
1) Perception: real-time video monitoring and interaction capturing.
2) Decision: raising proactive interaction in proper situations,
3) Reaction: continuous interaction with users.
However, inherent conflicts exist among the desired capabilities. 
The Decision and Reaction require a contrary Perception scale and grain, and the autoregressive decoding blocks the real-time Perception and Decision during the Reaction.
To unify the conflicted capabilities within a harmonious system, we present Dispider, a system that disentangles Perception, Decision, and Reaction.
Dispider features a lightweight proactive streaming video processing module that tracks the video stream and identifies optimal moments for interaction. Once the interaction is triggered, an asynchronous interaction module provides detailed responses, while the processing module continues to monitor the video in the meantime. Our disentangled and asynchronous design ensures timely, contextually accurate, and computationally efficient responses, making Dispider ideal for active real-time interaction for long-duration video streams.
Experiments show that Dispider not only maintains strong performance in conventional video QA tasks, but also significantly surpasses previous online models in streaming scenario responses, thereby validating the effectiveness of our architecture. The code and model are released at \url{https://github.com/Mark12Ding/Dispider}. 

\end{abstract}    
\section{Introduction}
\label{sec:intro}

The recent surge in multimodal large language models (MLLMs) has brought considerable attention to video understanding tasks. Given a video, these models aim to accurately comprehend its content and provide responses aligned with human expectations. However, the majority of current video LLMs are designed around an offline setting, where models are required to view the entire video before generating a single response. While effective for certain applications, this offline approach is impractical for real-time, interactive scenarios where users expect continuous feedback as video streams. 

\begin{figure}
    \centering
    \includegraphics[width=\linewidth]{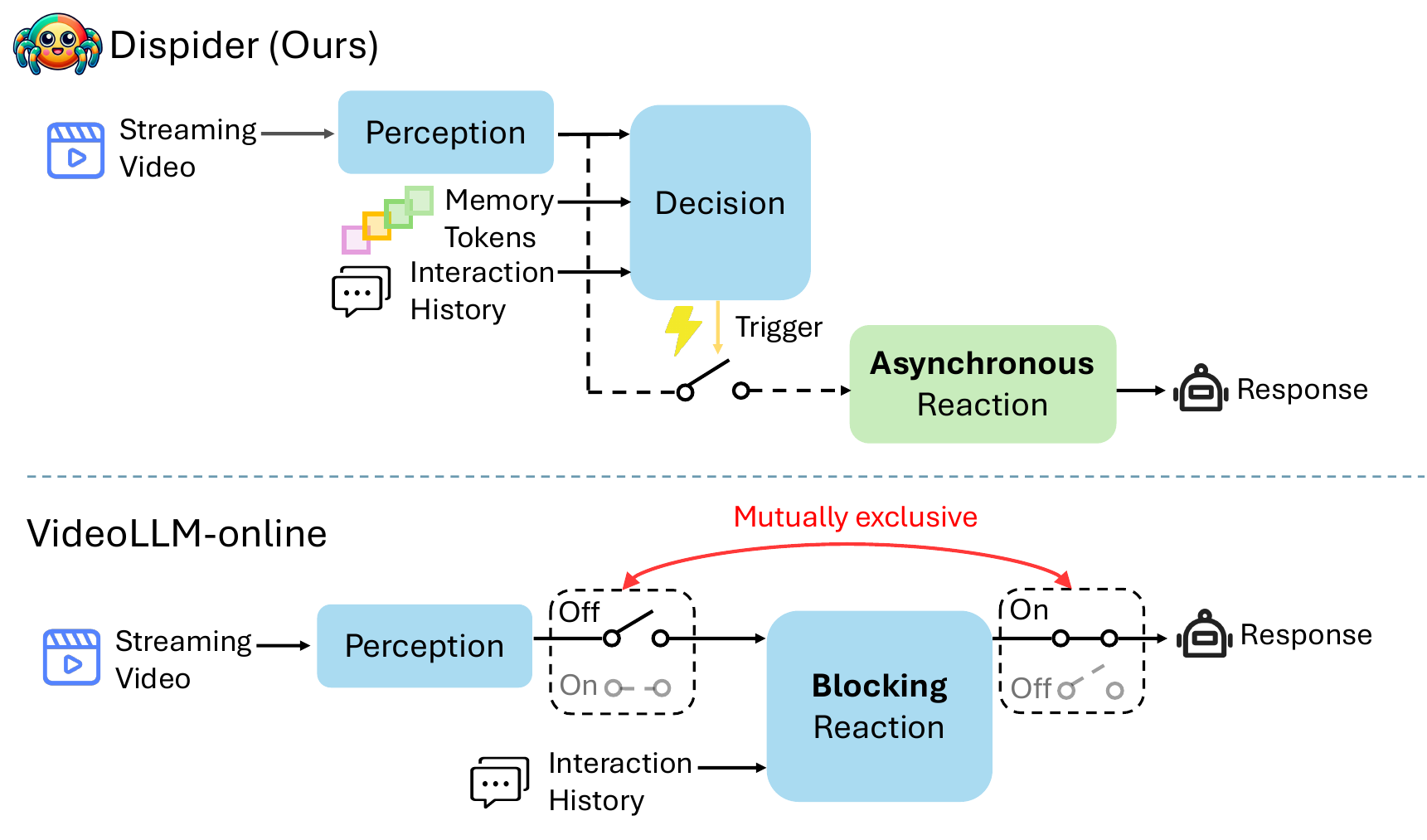}
    \caption{Schematic comparison between Dispider and VideoLLM-online~\cite{videollm-online}. Our Dispider introduces a disentangled paradigm for active real-time interaction with streaming video. It features a lightweight perception module for continuous monitoring, a decision module to determine when to trigger system interactions, and an asynchronous reaction module for generating detailed responses. In contrast, VideoLLM-online is unable to simultaneously perform streaming perception and response generation because it relies on a single LLM to handle both functions.}
    \label{fig:comp}
\end{figure}

VideoLLM-online~\cite{videollm-online} has been a pioneering effort in addressing this issue. By leveraging the LLaMA model~\cite{li2023llama}, they enable proactive response updates during streaming video inputs. However, it has a critical limitation: since it uses a single LLM for video processing and response generation, it cannot perform perception and answer reaction simultaneously. The autoregressive nature of next-token prediction in LLMs forces VideoLLM-online to alternate between perception (processing frames) and reaction (generating response), rather than handling both in parallel. This blocking reaction delays the video inputs during responding, reduces responsiveness and hinders its ability to engage in real-time interactions effectively, especially during long-duration video streams that require continuous, uninterrupted processing.

To overcome these limitations, we introduce Dispider, a novel system designed for active, real-time interaction with streaming videos. 
Dispider disentangles \textbf{perception}, \textbf{decision}, and \textbf{reaction} into asynchronous modules that operate in parallel. As illustrated in Figure~\ref{fig:comp}, the perception module continuously processes streaming video inputs. Meanwhile, the decision module integrates the full interaction history including previous decision tokens and visual information from the perception module. This determines whether to trigger the reaction module, thereby ensuring that detailed, timely responses can be delivered. Unlike VideoLLM-online, Dispider’s decision process remains uninterrupted by the asynchronous reaction step, ensuring a fluid and continuous decision-making flow.

Specifically, we design Dispider with a scene-based perception module, a real-time response decision module, and an asynchronous interaction module. In the scene-based perception module, the system dynamically segments the video stream into non-uniform clips based on scene boundaries, ensuring that each segment captures meaningful changes in the visual content. Subsequently, we integrate scene-based features, historical responses, and previous decision tokens into an interleaved sequence. This sequence is then processed by the real-time response decision module, which determines whether the model should generate a response or continue waiting for additional video content.

When an interaction is triggered, the asynchronous interaction module generates context-aware responses without disrupting the ongoing video processing. This asynchronous approach ensures that video analysis and response generation occur in parallel, maintaining the system's real-time performance.

Furthermore, Dispider is trained on a specialized streaming QA dataset designed to simulate real-time interaction scenarios. This training recipe enables the model to effectively handle both instances requiring responses and situations where silence is more appropriate. Consequently, Dispider enhances its ability to interact appropriately across diverse contexts, ensuring timely and relevant responses in dynamic streaming environments.

This disentangled design ensures that the Dispider can provide timely, accurate, and computationally efficient responses, even for long-duration video streams. We evaluate Dispider's performance in real-time video stream interactions (StreamingBench~\cite{lin2024streaming}) and show that it significantly outperforms VideoLLM-online~\cite{videollm-online} in temporal grounding, proactive response generation, and multi-step reasoning. Furthermore, it outperforms conventional offline Video LLMs across long-video benchmarks (EgoSchema~\cite{mangalam2024egoschema}, VideoMME~\cite{fu2024video}, MLVU~\cite{zhou2024mlvu}) and time-sensitive tasks (ETBench~\cite{liu2024etbench}). Dispider demonstrated exceptional performance, particularly excelling in temporal reasoning and effectively handling diverse video lengths.

\section{Related Work}
\label{sec:rw}

\subsection{Large Language Models}
Large Language Models (LLMs) have revolutionized natural language processing, achieving remarkable performance across a wide range of tasks. Early models such as BERT~\cite{devlin2018bert} and T5~\cite{raffel2020exploring} used masked language modeling for pre-training. The shift to decoder-only models like GPT~\cite{radford2018improving} introduced scalable architectures that enhanced language generation. Recent models, including PaLM~\cite{chowdhery2023palm}, LLaMA~\cite{touvron2023llama}, and GPT-4~\cite{openai2023gpt4}, continue to push the boundaries with massive parameters and extensive training data. Techniques such as supervised fine-tuning and reinforcement learning from human feedback~\cite{ouyang2022training, bai2022training, cai2024internlm2, bai2023qwen, young2024yi, li2023llama, team2023gemini, team2024gemini, vicuna2023, jiang2023mistral} have further improved these models' ability to generate coherent, contextually relevant responses. Inspired by the reasoning capabilities of LLMs, we extend these models to the domain of streaming video understanding, where real-time interaction presents unique challenges.

\subsection{Video Large Language Models}
The progress of multi-modal LLMs~\cite{li2023blip2bl, Qwen2VL, liu2023llava, zhang2023internlm,dong2024internlm4khd, xing2024pyramiddrop, Qi_2024_CVPR} has been significant, particularly with image-based models like InstructBLIP~\cite{instructblip}, LLaVA~\cite{liu2023improved,liu2024llavanext,liu2023llava}, and Flamingo~\cite{alayrac2022flamingo}, which integrate vision and language models. Extending this to video introduces challenges in managing both frame sequences and the context length of LLMs. Recent works on video LLMs~\cite{li2024llamavid, wang2024videotree, fan2024videoagent, maaz2023videochatgpt, zhang2023simple,liu2024kangaroo,weng2025longvlm,song2024moviechat, lin2023mm,wu2024motionllm,chen2024motionllm,ye2024mplug,xu2023mplug2,chen2024sharegpt4video,chen2023internvl,zhang2024internlm, damonlpsg2023videollama, cheng2024videollama, fu2024vita, fu2023mme, dong2024internlm, zhang2024internlmomni} present novel strategies for processing long-form videos while maintaining effective reasoning. Notable approaches include TimeChat~\cite{ren2024timechat}, which emphasizes temporal relationships across video frames, and MovieChat~\cite{song2024moviechat}, which uses sparse memory to handle long videos. VideoChat~\cite{li2023videochat} adopts a chat-centric approach, integrating video foundation models with LLMs to excel at spatiotemporal reasoning, event localization, and causal inference. Building on these video LLMs, our work introduces a pipeline designed to process streaming video inputs and generate real-time outputs.

\subsection{Streaming Video Understanding}
In practice, users typically expect models to operate online and interactively in real-time. This is known as streaming video understanding, which involves processing continuous video streams while ensuring long-term temporal consistency and enabling interactive responses. Despite its practical significance, only a few works have explored this area.
VideoLLM-online~\cite{videollm-online} introduces the LIVE framework for streaming dialogue, but it lacks an efficient module for handling long-term video inputs over extended periods and does not prioritize real-time interactivity. VideoStream~\cite{qian2024streaming} proposes memory-propagated encoding for long videos, yet its focus is on offline processing rather than real-time streaming. Similarly, Flash-VStream~\cite{flashvstream} addresses inference latency and memory efficiency for long video streams but overlooks real-time user interaction.
In contrast, our approach centers on developing a real-time visual assistant for streaming video, emphasizing both long-context handling and interactive, real-time responses, thereby bridging the gap left by previous methods.

\section{Method}
\subsection{Problem Formulation}
In contrast to previous offline video LLMs, which generate responses only after processing the entire video, our approach operates in real-time by simultaneously processing the video and providing continuous, interactive responses. Specifically, the model actively determines when sufficient information has been observed to provide a complete response, allowing it to produce answers promptly without waiting for the entire video to finish. Following the formulation of VideoLLM-online~\cite{videollm-online}, we formulate this new setting as below.

Given a continuous video stream \( V = \{v_1, v_2, \ldots, v_T\} \), where \(v_i\) represents the $i$-th video clip, and a context sequence \(C_t\) up to a specific time \(t\) (which includes prior vision-language interactions such as historical user queries and assistant responses), our goal is to generate timely and accurate dialogue responses \(R_t\) without processing the entire video sequence.

At each time \(t\), the model observes \(C_t\) and the video frames up to that time \(V_{[0, t]} = \{v_1, v_2, \ldots, v_t\}\). The model must decide whether it has sufficient information to generate a response. We define a decision function \(\pi\) and a reaction function \(f\):

\[
a_t = \pi(C_t, V_{[0, t]}) \in \{\text{wait}, \text{respond}\}.
\]

If \(a_t = \text{respond}\), the model generates a response:

\[
R_t = f(C_t, V_{[0, t]})
\]
Otherwise, the model keeps silent for more information.

In this work, we implement these functions using a disentangled framework composed of three distinct modules: Perception, Decision, and Reaction. These modules are designed to handle the unique challenges of real-time video understanding and dialogue generation. The Perception module focuses on continuous video monitoring, while the Decision module assesses when to act, and the Reaction module generates responses without waiting for the entire video sequence to finish. This approach allows the system to remain responsive to new information while ensuring that the generated answers are based on the most relevant and up-to-date context.

\begin{figure*}
    \centering
    \includegraphics[width=\linewidth]{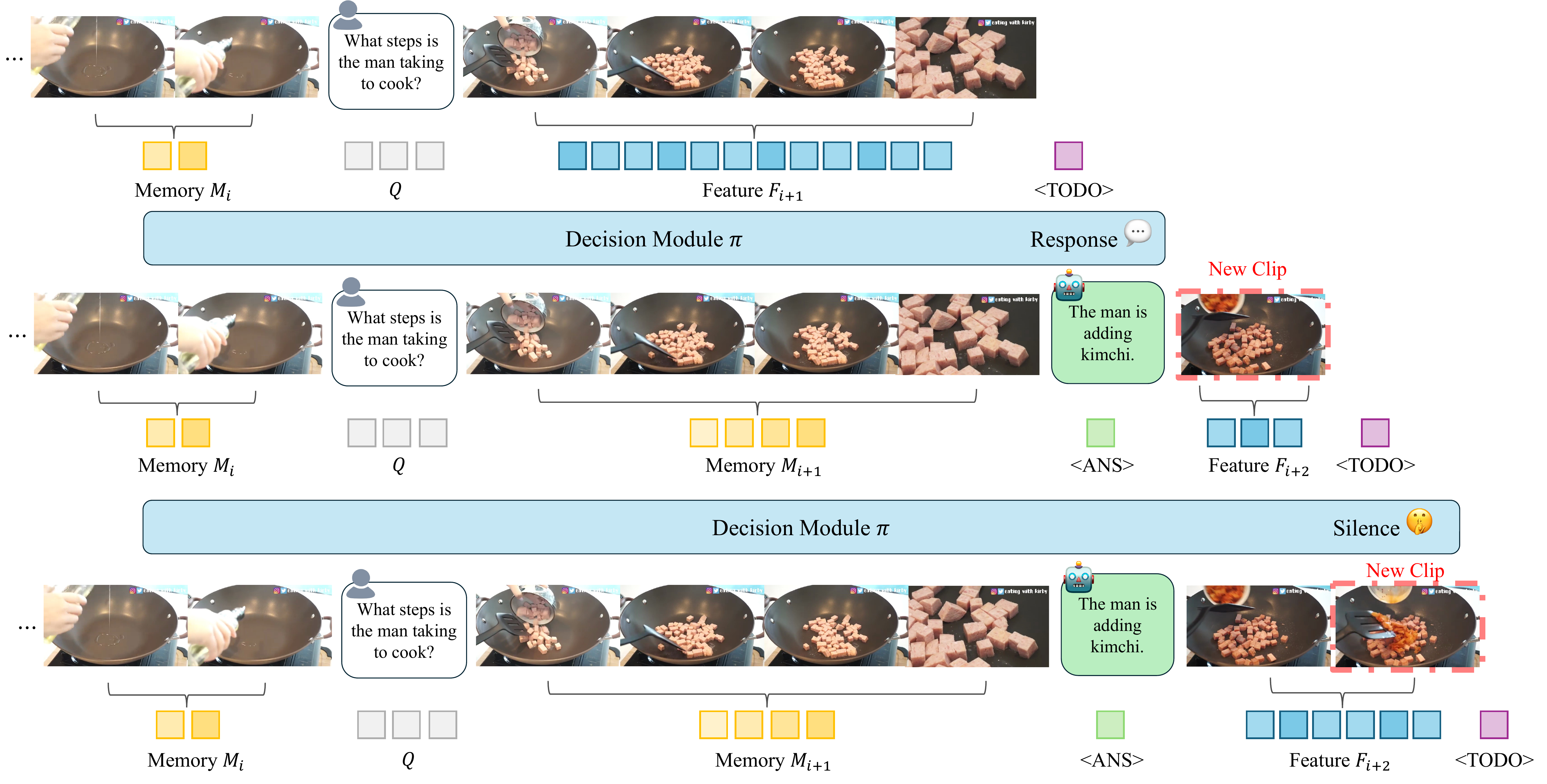}
    \caption{ Illustration of the Response Decision Module in a proactive streaming video processing system. The module dynamically determines when to respond during video streaming by segmenting the video into non-uniform clips and utilizing historical memory to capture context. The module then combines memory features, clip features, question text, and special tokens, ⟨TODO⟩ and ⟨ANS⟩, to decide on response timing.}
    \label{fig:pipeline}
\end{figure*}

\subsection{Proactive Streaming Video Processing}
To enable active real-time response, we propose a proactive streaming video processing approach that dynamically segments the video stream and evaluates whether to generate a response. 
The system is composed of two key modules: the Scene-based Perception Module and the Real-time Response Decision Module.

\vspace{2pt}
\noindent\textbf{Scene-based Perception Module.}
To ensure efficient processing of long video streams, we begin by adaptively segmenting the video into non-uniform clips based on scene boundaries. This segmentation strategy preserves the structural information of the video, allowing the model to focus on the most informative parts while removing redundancy and maintaining context.

We begin by sampling frames at a regular interval and extract L2-normalized feature embeddings using the pre-trained SigLip model~\cite{zhai2023sigmoid}. By computing the cosine similarity between these embeddings, we can identify significant visual changes, which indicate potential scene boundaries. These boundaries help divide the video into meaningful segments. To prevent excessively short clips, we introduce an exclusion window around the boundaries, ensuring that the resulting clips are of sufficient length to contain relevant information.

Each clip $v_i$ is then processed by the video encoder to produce the clip-wise feature representation $F_i$, along with a special clip indicator~$\hat{F}_i$. These clip features are used in the subsequent decision-making process to determine if enough information has been gathered to respond.

\vspace{2pt}
\noindent\textbf{Real-time Response Decision Module.}
Based on the video content observed so far and the historical context, the real-time response decision module evaluates whether the model should generate a response or continue waiting for more video content. We illustrate this whole process in Figure~\ref{fig:pipeline}. To effectively combine these multi-modal inputs, we use an interleaved sequence format, which integrates video features, question information, and decision tokens. 

The process begins with the sequence of video clip representations up to the point of the user’s question, followed by the question $Q$ and a special token, $\langle \text{TODO} \rangle$, which indicates that a response decision is yet to be made. Formally, the input sequence at this stage is:
$$F[0: q_{\text{pos}}]\oplus Q\oplus \langle \text{TODO} \rangle,$$ 
where $q_{\text{pos}}$  is the timestamp of the question, and $\oplus$ denotes concatenation.

After this initial stage, the model aggregates the features up to the question timestamp into a global memory $M$ through pooling. Then, we extend the sequence to include this global memory that captures historical context, the question texts, and the clip features from the question timestamp to the current point, ending with $\langle \text{TODO} \rangle$: 
$$M[0: q_{\text{pos}}]\oplus Q\oplus F[q_{\text{pos}}:]\oplus \langle \text{TODO} \rangle.$$

When the model decides to respond, the $\langle \text{ANS} \rangle$ token is used to mark the response action. Specifically, with the video continuing to stream, we further integrate global memory from the question timestamp to the position of the first answer, insert the special answer token $\langle \text{ANS} \rangle$, and include later clips for potential subsequent responses. This sequence concludes with the current clip memory segment and a $\langle \text{TODO} \rangle$ token. Thus, the ultimate input format with multiple answers can be written as:
\begin{align*}
&M[0: q_{\text{pos}}]  \oplus   Q  \oplus \\
&M[q_{\text{pos}}: a^1_{\text{pos}}]  \oplus \langle \text{ANS} \rangle \oplus \\
&M[a^1_{\text{pos}}: a^2_{\text{pos}}]  \oplus  \langle \text{ANS} \rangle \oplus \\
&\cdots \\
&F[a^k_{\text{pos}}:] \oplus \langle \text{TODO} \rangle,
\end{align*}
where $a^k_{\text{pos}}$ stands for the timestamp for the $k$-th answer. 

Importantly, none of the tokens we utilize originate from the responses generated by the Reaction module. This design ensures that the Decision module remains unblocked by the response generation process, allowing it to continuously monitor the video stream.

We feed the interleaved sequence into a compact LLM and adopt a binary classification head on top of the final-layer embedding of the $\langle \text{TODO} \rangle$ token. This head is trained to predict whether the model should respond at each timestamp. We use standard binary cross-entropy loss to supervise the model's decision-making process.

\subsection{Asynchronous Interaction}
Once the interaction is triggered, a dedicated asynchronous module is employed to generate contextually fine-grained responses tailored to the state of the video stream. Specifically, for the $k$-th answer output, the interaction module receives the current query, the $k-1$ previously produced answers, and the grounded clip features extracted at the corresponding timestamp. Relevant historical clips are retrieved by computing their cosine similarity with the embedding of a designated $\langle \text{TODO} \rangle$ token. This procedure supports multi-hop reasoning, where relevant evidence may be distributed across multiple temporal segments.

To train the model’s temporal multi-hop retrieval capabilities, we compute the cosine similarity between the $\langle \text{TODO} \rangle$ token embedding and each historical clip indicator $\hat{F}_i$. Applying the softmax function to these similarity scores yields a predicted relevance distribution \( \hat{P}_{\text{pred}}(i) \), representing the relevance of $i$-th historical clip to the current response.

For supervision, the ground-truth temporal relevance is represented by a binary mask over clips known to be relevant to the current question $Q$. Defining $R_Q$ as the set of ground-truth relevant clips, we construct the true relevance distribution as:
\[
\hat{P}_{\text{true}}(i) = \frac{1}{|R_Q|} \quad \text{if} \quad i \in R_Q  \quad \text{else} \quad 0,
\]
where \( R_Q \) is the ground-truth set of relevant clips for particular question \( Q \).

We then calculate the KL divergence loss between the true distribution, \( \hat{P}_{\text{true}} \), and the predicted distribution, \( \hat{P}_{\text{pred}} \), as:
\[
\mathcal{L}_{\text{KL}} = -\frac{1}{|R_Q|} \sum_{i=1}^{|R_Q|} \hat{P}_{\text{true}}(i) \log \hat{P}_{\text{pred}}(i).
\]
This KL divergence loss encourages the model to align its predictions with the true temporal relevance of the clips, thereby improving its ability to correctly retrieve and ground the relevant video segments for generating accurate, timely responses.

In addition, the response decision module may occasionally trigger the interaction actually when no response is required. To accommodate this scenario, we introduce both positive (response-required) and negative (no-response-required) samples when training the interaction module, enabling the model to simulate real-time interactive conditions. The model thus learns to either generate incremental, contextually enriched responses based on newly emerging cues in the video stream or produce a special $\langle \text{SILENT} \rangle$ token to indicate silence when appropriate.

This adaptive reasoning approach ensures that the model remains responsive to new information. By focusing on unanswered content and utilizing past interactions, the model delivers timely and relevant responses, enhancing the overall efficiency and user experience in streaming dialogue generation.

In this way, we decouple the response generation from the video stream processing. This separation allows the streaming video to continue being processed in parallel, without waiting for response generation to complete. As a result, the system remains highly efficient and responsive, ensuring that the video content is continuously monitored and processed while still generating contextually accurate responses at the appropriate moments.

\section{Experiments}
\subsection{Implementation Details} 
Dispider utilizes a compact LLM as the proactive streaming video processor for response decisions, and a larger LLM for the precision interaction module. In practice, the input video frames are resized to $224\times 224$, and a CLIP-L/14~\cite{radford2021learning} is employed to extract frame-wise features. Following the token compression techniques in VideoStream~\cite{qian2024streaming}, we concatenate adjacent tokens and use the compact LLM, instantiated as Qwen2-1.5B~\cite{qwen2}, to produce time-aware compressed clip-wise features along with clip indicators. Next, we reuse this compact LLM to process the sequence consisting of global memory, question texts, and clip features for response decisions. The final LLM, instantiated as Qwen2-7B~\cite{qwen2}, receives the grounded clips and global memory to generate responses at the necessary timestamps.

We adopt a two-stage training process. In the first stage, we train the streaming video processor and response decision module using a combination of GroundVQA~\cite{di2024grounded} and ET-Instruct~\cite{liu2024etbench} with enriched temporal annotations for supervising streaming responses and providing temporal grounding labels. To further enhance basic reasoning capabilities, we add 50K curated implicit QA pairs with time labels. Next, we construct a dataset of 122K streaming video QA pairs, derived from timestamped QA in ET-Instruct~\cite{liu2024etbench} and augmented with data from VideoChatGPT~\cite{maaz2023videochatgpt} and LLaVA-Next-Video~\cite{li2024llava}, to train the reaction module.

In the second training stage, we freeze the video encoder and the compact LLM, then train only the final interaction module. During training, we insert instructions at various timestamps to improve adaptability. For inference on conventional benchmarks, the question is placed at the end of the video to ensure fair comparisons with prior work, whereas for streaming evaluation, it is posed at the beginning to enable proactive responses.

\subsection{Benchmarks}
For evaluation, we utilize a range of benchmarks tailored to different aspects of long-video QA and streaming video understanding.

\vspace{2pt}
\noindent\textbf{StreamingBench.} StreamingBench~\cite{lin2024streaming} serves as the latest comprehensive benchmark for evaluating streaming video understanding in multimodal large language models (MLLMs). It emphasizes three critical evaluation aspects: Real-time Visual Understanding, Omni-source Understanding, and Contextual Understanding. The benchmark includes a diverse dataset of 900 videos paired with 4,500 human-annotated QA pairs, with five questions per video asked at varying timestamps.

\vspace{2pt}
\noindent\textbf{ETBench Subset.} In addition to StreamingBench, we construct a streaming video QA benchmark using a subset of ETBench to measure the model's proactive response capability in real-time video interactions. Specifically, we select six subtasks from ETBench suitable that require the model to predict explicit event timestamps: step localization (SLC), dense video captioning (DVC), temporal action localization (TAL), temporal video grounding (TVG), episodic memory (EPM), and video highlight detection (VHD), encompassing a total of 4,460 videos. For this benchmark, we test our model using two versions. In the conventional setting, we pose the instruction at the end of the question. In the streaming setting, the instruction is provided at the beginning of each video, and the model is required to produce the correct responses at appropriate timestamps as the video plays. We report the F1 score for temporal grounding evaluation and the sentence similarity score for caption evaluation.

\vspace{2pt}
\noindent\textbf{Long-Video QA Benchmarks.} Finally, we adopt several long-video QA benchmarks, including EgoSchema~\cite{mangalam2024egoschema}, VideoMME~\cite{fu2024video}, and MLVU~\cite{zhou2024mlvu}. EgoSchema consists of over 5K videos, each approximately 3 minutes long, while VideoMME and MLVU include videos of varying lengths, from a few minutes to several hours. We report accuracy on multiple-choice questions for comparison across these benchmarks.

\subsection{Streaming Video Understanding}
We evaluate Dispider’s performance in streaming video interactions, emphasizing its ability to process real-time input and respond dynamically. Questions are posed at the start of the video, and the model generates responses only when relevant clues are detected, remaining silent otherwise for meaningful, context-aware interactions.

As shown in Table~\ref{tab:streamingbench}, Dispider significantly outperforms Flash-VStream~\cite{flashvstream} and VideoLLM-online~\cite{videollm-online}, particularly in the Proactive Output (PO) task. This task requires the model to determine the precise timing of its response, such as outputting ``GOAL" when a goal is scored. Unlike standard input-output tasks, it demands proactive maintaining an internal state to track relevant video frames.

While other streaming models fail in this task, Dispider achieves a competitive score of 25.3. Even compared to offline Video LLMs, where the question is provided after the video has played, Dispider demonstrates superior proactive decision-making capabilities by handling questions posed at the start of the video.

For the ETBench subset in the streaming setting, as shown in the bottom rows of Table~\ref{etbenchexp}, our model consistently outperforms VideoLLM-Online~\cite{videollm-online} across all metrics, with particularly notable improvements in temporal grounding.
This demonstrates that our disentangled design equips the model with much stronger temporal awareness, enabling more proactive responses aligned with specific instructions.
Notably, on dense video captioning and step localization tasks, our model achieves both more precise temporal grounding and more comprehensive descriptions in streaming mode than the conventional setting. This indicates that our disentangled architecture can effectively monitor the video stream in real-time and proactively generate informative responses according to the instructions. And the streaming interaction has the potential to achieve more powerful video understanding.

\begin{figure*}
    \centering
    \includegraphics[width=\linewidth]{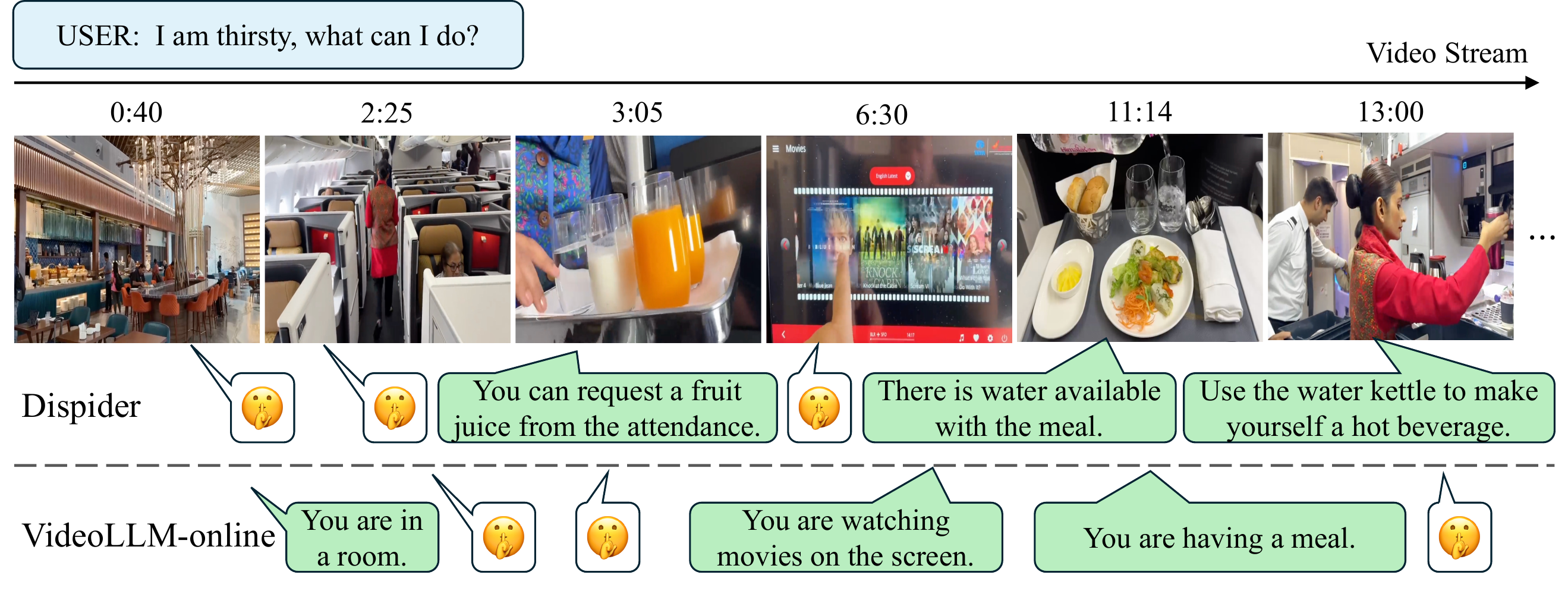}
    \caption{Quantitative comparison between Dispider and VideoLLM-online in streaming video understanding. The question is inserted at the beginning, and we show the model's response in either answer texts or the state of keeping silent.}
    \label{visualize}
\end{figure*}

Additionally, we present a quantitative comparison between our model and VideoLLM-online in Fig.~\ref{visualize}. 
From this comparison, when faced with questions requiring multi-step reasoning, our model can gradually identify the necessary clues from the video stream and generate informative answers step by step, while VideoLLM-online~\cite{videollm-online} fails to do so. For example, from the `thirsty' in the question, our model can associate it with the drinks appearing in the video stream, and infer what actions are needed based on the context. In contrast, VideoLLM-online only gives simple descriptions of the scene or ongoing actions.

A key advantage of our method is the disentangled perception, decision, and reaction architecture, which enables the model to simultaneously process the streaming video input and generate responses in a non-blocking manner. However, VideoLLM-online has to experience interruptions in the video stream during answer generation as shown in Figure~\ref{fig:pipeline}.

\begin{table*}[t]
\centering
\renewcommand{\arraystretch}{1.4}
\Huge
\fontseries{b}
\setlength{\arrayrulewidth}{1pt}
\begin{adjustbox}{max width=\textwidth}
\begin{tabular}{lcc|ccccccccccc|ccccc|ccccc|ccc|c}
\toprule
\multirow{2}{*}{\textbf{Model}} & \multirow{2}{*}{\textbf{Params}} & \multirow{2}{*}{\textbf{Frames}} & \multicolumn{11}{c|}{\textbf{Real-Time Visual Understanding}} & \multicolumn{5}{c|}{\textbf{Omni-Source Understanding}} & \multicolumn{5}{c|}{\textbf{Contextual Understanding}} & \multirow{2}{*}{\textbf{Overall}} \\
\cmidrule(lr){4-14} \cmidrule(lr){15-19} \cmidrule(lr){20-24}
 &  &  & OP & CR & CS & ATP & EU & TR & PR & SU & ACP & CT & \textbf{All}  & ER & SCU & SD & MA & \textbf{All} & ACU & MCU & SQA & PO & \textbf{All} \\\hline
\multicolumn{25}{c}{\textbf{Human}} \\\hline
Human{$^\ddag$} & - & - & 89.47 & 92.00 & 93.60 & 91.47 & 95.65 & 92.52 & 88.00 & 88.75 & 89.74 & 91.30 & 91.46  & 88.00 & 88.24 & 93.60 & 90.27 & 90.26  & 88.80 & 90.40 & 95.00 & 100 & 93.55 & 91.66\\\hline
\multicolumn{25}{c}{\textbf{Proprietary MLLMs}} \\\hline
Gemini 1.5 pro & - & 1 fps & 79.02 & 80.47 & 83.54 & 79.67 & 80.00 & 84.74 & 77.78 & 64.23 & 71.95 & 48.70 & 75.69 & 46.80 & 39.60 & 74.90 & 80.00 & 60.22 & 51.41  & 40.73 & 54.80 & 45.10 & 48.73 & 67.07 \\
GPT-4o & - & 64 & 77.11 & 80.47 & 83.91 & 76.47 & 70.19 & 83.80 & 66.67 & 62.19 & 69.12 & 49.22 & 73.28 & 41.20 & 37.20 & 43.60 & 56.00 & 44.50  & 41.20 & 38.40 & 32.80 & 56.86 & 38.70 &60.15\\
Claude 3.5 Sonnet & - & 20 & 80.49 & 77.34 & 82.02 & 81.73 & 72.33 & 75.39 & 61.11 & 61.79 & 69.32 & 43.09 & 72.44 & 31.60 & 34.00 & 32.80 & 48.80 & 36.80  & 38.40 & 34.80 & 34.40 & 64.71 & 37.70 & 57.68\\\hline
\multicolumn{25}{c}{\textbf{Open-Source Video MLLMs}} \\\hline
LLaVA-OneVision & 7B & 32 & 80.38 & 74.22 & 76.03 & 80.72 & 72.67 & 71.65 & 67.59 & 65.45 & 65.72 & 45.08 & 71.12 & 40.80 & 37.20 & 33.60 & 44.80 & 38.40  & 35.60 & 36.00 & 27.27 & 29.55 & 32.74 & 56.36\\
Qwen2-VL & 7B & 0.2-1 fps & 75.20 & 82.81 & 73.19 & 77.45 & 68.32 & 71.03 & 72.22 & 61.19 & 61.47 & 46.11 & 69.04 & 41.20 & 22.00 & 32.80 & 43.60 & 34.90  & 31.20 & 26.00 & 39.60 & 22.73 &  31.66& 54.14\\
MiniCPM-V 2.6 & 8B & 32 & 71.93 & 71.09 & 77.92 & 75.82 & 64.60 & 65.73 & 70.37 & 56.10 & 62.32 & 53.37 & 67.44& 40.80 & 24.00 & 34.00 & 41.20 & 35.00  & 34.00 & 31.60 & 41.92 & 22.22 & 34.97 & 53.85\\
LLaVA-NeXT-Video & 32B & 64 & 78.20 & 70.31 & 73.82 & 76.80 & 63.35 & 69.78 & 57.41 & 56.10 & 64.31 & 38.86 & 66.96 & 37.69 & 24.80 & 34.40 & 42.80 & 34.90  & 29.20 & 30.40 & 35.35& 18.18 & 30.79 & 52.77\\
InternVL-V2 & 8B & 16 & 68.12 & 60.94 & 69.40 & 77.12 & 67.70 & 62.93 & 59.26 & 53.25 & 54.96 & 56.48 & 63.72& 37.60 & 26.40 & 37.20 & 42.00 & 35.80 & 32.00 & 31.20 & 32.32 & 40.91 & 32.42 & 51.40\\
Kangaroo & 7B & 64 & 71.12 & 84.38 & 70.66 & 73.20 & 67.08 & 61.68 & 56.48 & 55.69 & 62.04 & 38.86 & 64.60 & 37.60 & 31.20 & 28.80 & 39.20 & 34.20 & 32.80 & 26.40 & 33.84 &  16.00& 30.06 & 51.10\\
LongVA & 7B & 128 & 70.03 & 63.28 & 61.20 & 70.92 & 62.73 & 59.50 & 61.11 & 53.66 & 54.67 & 34.72 & 59.96 & 39.60 & 32.40 & 28.00 & 41.60 & 35.40  & 32.80 & 29.60 & 30.30 & 15.91 & 29.95& 48.66\\
VILA-1.5 & 8B & 14 & 53.68 & 49.22 & 70.98 & 56.86 & 53.42 & 53.89 & 54.63 & 48.78 & 50.14 & 17.62 & 52.32& 41.60 & 26.40 & 28.40 & 36.00 & 33.10 & 26.80 & 34.00 & 23.23 & 17.65 & 27.35&43.20\\
Video-CCAM & 14B & 96 & 56.40 & 57.81 & 65.30 & 62.75 & 64.60 & 51.40 & 42.59 & 47.97 & 49.58 & 31.61 & 53.96 & 33.60 & 22.00 & 28.40 & 34.80 & 29.70 & 27.60 & 24.40 & 16.67 &  22.73& 22.88 &42.53 \\
Video-LLaMA2 & 7B & 32 & 55.86 & 55.47 & 57.41 & 58.17 & 52.80 & 43.61 & 39.81 & 42.68 & 45.61 & 35.23 & 49.52 & 30.40 & 32.40 & 30.40 & 36.00 & 32.40  & 24.80 & 26.80 & 18.67 & 0.00 & 21.93 & 40.40\\\hline
\multicolumn{25}{c}{\textbf{Streaming MLLMs}} \\\hline
Flash-VStream & 7B& -& 25.89& 43.57 &24.91 &23.87& 27.33& 13.08 &18.52& 25.20& 23.87 &48.70 &23.23  &25.91& 24.90& 25.60& 28.40& 26.00& 24.80& 25.20 &26.80 &1.96& 24.12 & 24.04\\
VideoLLM-online & 8B & 2 fps &  39.07 &40.06 &34.49& 31.05& 45.96 &32.40 &31.48& 34.16 &42.49 &27.89& 35.99 &31.20 &26.51 &24.10 &32.00 &28.45 &24.19 &29.20 &30.80 &3.92 &26.55 & 32.48\\
\textbf{Dispider (ours)}  & 7B & 1 fps & 74.92& 75.53& 74.10& 73.08& 74.44& 59.92& 76.14& 62.91& 62.16& 45.80&67.63 &35.46 &25.26& 38.57 &43.34 &35.66& 39.62& 27.65& 34.80& 25.34 &33.61 & 53.12\\\bottomrule
\end{tabular}
\end{adjustbox}
\caption{Performance comparison on StreamingBench on Omni-source Understanding, Contextual Understanding, and Real-Time Visual Understanding. Omni-source Understanding includes Emotion Recognition (ER), Scene Understanding (SCU), Source Discrimination (SD), and Multimodal Alignment (MA). Contextual Understanding includes Misleading Context Understanding (MCU), Anomaly Context Understanding (ACU), Sequential Question Answering (SQA) and Proactive Output (PO). Real-Time Visual Understanding includes Object Perception (OP), Causal Reasoning (CR), Clips Summarization (CS), Attribute Perception (ATP), Event Understanding (EU), Text-Rich Understanding (TR), Prospective Reasoning (PR), Spatial Understanding (SU), Action Perception (ACP), and Counting (CT). Results are categorized into Human, Proprietary MLLMs, and Open-Source MLLMs for a comprehensive evaluation.}
\label{tab:streamingbench}
\end{table*}

\subsection{Conventional Video Understanding}
In this section, we compare our model with existing video LLMs on conventional video QA benchmarks, where the model is required to provide one answer after watching the complete video.

We first present a comparative analysis across three challenging long-video benchmarks. As shown in Table~\ref{longexp}, we report the accuracy on the EgoSchema full set, MLVU multiple-choice questions, and the VideoMME overall set without subtitles. Generally, our disentangled architecture handles this conventional scenario well and achieves competitive performance. Notably, on EgoSchema, which requires long temporal reasoning, our method achieves a leading performance of 55.6, demonstrating strong temporal perception.

In terms of MLVU and VideoMME, which consist of videos ranging from minutes to hours long, our model’s promising results highlight its ability to efficiently process contextual information across varying temporal lengths. This capability is crucial for streaming video interactions, as it allows the model to accurately summarize context when questions are inserted at arbitrary temporal positions.

Additionally, we adopt a recent time-aware benchmark, ETBench~\cite{liu2024etbench}, to evaluate temporal awareness. In the conventional setting in Table~\ref{etbenchexp}, our method is able to capture the timestamps corresponding to specific questions. Without a specialized design for time tokens, our model achieves the highest F1 score in temporal video grounding and episodic memory subtasks, even surpassing models with specialized temporal tokens. The promising performance on dense video captioning and step localization further demonstrates the model's ability in both accurate temporal grounding and precise visual perception.

\begin{table*}
    \centering
    \small
    \begin{tabular}{lccccc}
    \toprule
        Method & LLM Size & Frames & EgoSchema & MLVU & VideoMME \\
        \midrule
        Video-LLaVA~\cite{lin2023video} & 7B & 8 & 38.4 & 47.3 & 39.9 \\
        Chat-UniVi~\cite{jin2023chatunivi} & 7B & 64 & - & - & 40.6 \\
        LLaMA-VID~\cite{li2024llamavid} & 7B & 1 FPS & 38.5 & 33.2 & - \\
        TimeChat~\cite{ren2024timechat} & 7B & 96 & 33.0 & 30.9 & 30.2 \\
        MovieChat~\cite{song2024moviechat} & 7B & 2048 & 53.5 & 25.8 & 38.2 \\
        Video-LLaMA2~\cite{cheng2024videollama} & 7B & 16 & 51.7 & 48.5 & 47.9 \\
        LLaVA-Next-Video~\cite{zhang2024llavanextvideo} & 7B & 32 & 43.9 & - & 46.6 \\
        ShareGPT4Video~\cite{chen2024sharegpt4video} & 8B & 16 & - & 46.4 & 39.9 \\
        VideoChat2~\cite{li2024mvbench} & 7B & 16 & 54.4 & 47.9 & 39.5 \\
        LongVA~\cite{zhang2024longva} & 7B & 128 & - & 56.3 & 52.6 \\
        Kangaroo~\cite{liu2024kangaroo} & 8B & 64 & - & 61.0 & 56.0 \\
        Video-CCAM~\cite{fei2024video} & 14B & 96 & - & 63.1 & 53.2 \\
        VideoXL~\cite{shu2024video} & 7B & 128 & - & 64.9 & 55.5 \\
        \textbf{Dispider (ours)} & 7B & 1 FPS & 55.6 & 61.7 & 57.2 \\ 
        \bottomrule
    \end{tabular}
    \caption{Comparison on long video benchmarks. We report the accuracy on the EgoSchema full set, MLVU multiple-choice questions, and the VideoMME overall set without subtitles. For a fair comparison, we also present the model size of the LLM and the number of sampled frames.}
    \label{longexp}
\end{table*}

\begin{table*}
    \centering
    \small
    \begin{tabular}{lccccccccc}
    \toprule
        Method & Frames & TVG$_{F1}$ & EPM$_{F1}$ & TAL$_{F1}$ & VHD$_{F1}$ & DVC$_{F1}$ & DVC$_{Sim}$ & SLC$_{F1}$ & SLC$_{Sim}$ \\
        \midrule
        \multicolumn{10}{l}{\textcolor{gray}{\emph{Conventional video QA inference.}}} \\
        \midrule
        \multicolumn{10}{l}{\emph{w/ specialized time tokens}} \\
        VTG-LLM~\cite{guo2024vtg} & 96 & 15.9 & 3.7 & 14.4 & 48.2 & 40.2 & 18.6 & 20.8 & 14.4 \\
        LITA~\cite{huang2025lita} & 100 & 22.2 & 4.6 & 18.0 & 23.9 & 39.7 & 17.2 & 21.0 & 12.2 \\
        ETChat~\cite{liu2024etbench} & 1 FPS & 38.6 & 10.2 & 30.8 & 62.5 & 38.4 & 19.7 & 24.4 & 14.6 \\
        \hdashline
        \multicolumn{10}{l}{\emph{w/o specialized time tokens}} \\
        VideoChatGPT~\cite{maaz2023videochatgpt} & 100 & 7.0 & 1.3 & 15.1 & 28.8 & 8.8 & 11.3 & 5.7 & 10.2 \\
        Video-LLaVA~\cite{lin2023video} & 8 & 7.0 & 1.9 & 15.0 & 28.9 & 28.0 & 15.0 & 0.9 & 8.3 \\
        LLaMA-VID~\cite{li2024llamavid} & 1 FPS & 5.5 & 1.2 & 8.0 & 30.0 & 27.1 & 12.6 & 5.2 & 11.1 \\
        Video-LLaMA2~\cite{cheng2024videollama} & 8 & 0.1 & 0.0 & 0.0 & 1,5 & 0.6 & 14.5 & 0.0 & 15.2 \\
        PLLaVA~\cite{xu2024pllava} & 16 & 6.9 & 1.1 & 5.7 & 28.9 & 13.3 & 10.6 & 9.7 & 11.8 \\
        VTimeLLM~\cite{huang2024vtimellm} & 100 & 7.6 & 1.9 & 18.2 & 28.9 & 12.4 & 13.1 & 8.7 & 6.4 \\
        TimeChat~\cite{ren2024timechat} & 96 & 26.2 & 3.9 & 10.1 & 40.5 & 16.6 & 12.5 & 5.6 & 9.2 \\
        \textbf{Dispider (ours)} & 1 FPS & 43.6 & 17.2 & 29.9 & 51.5 & 31.6 & 17.8 & 14.1 & 11.7 \\
        \midrule
        \multicolumn{10}{l}{\textcolor{gray}{\emph{Streaming video QA inference.}}} \\
        \midrule
        VideoLLM-Online~\cite{videollm-online} & 2 FPS & 13.2 & 3.8 & 9.1 & 22.4 & 24.0 & 13.4 & 9.9 & 10.1 \\
        \textbf{Dispider (ours)} & 1 FPS & 36.1 & 15.5 & 27.3 & 54.2 & 33.8 & 18.9 & 18.8 & 12.4 \\
        \bottomrule
    \end{tabular}
    \caption{Comparison on ETBench. We present the results for two different settings. In the conventional video QA setting, the model is required to answer the question after watching the entire video. In the streaming setting, the question is placed at the beginning of the video, and the model is expected to provide real-time responses. We report performance on six subtasks that are suitable for both evaluation settings.}
    \label{etbenchexp}
\end{table*}

\subsection{Ablation Study}

\noindent\textbf{Clip Segmentation Strategy.}
We first present an ablation study on clip segmentation. We compare the standard uniform clip segmentation (16 frames per clip sampled at 1 FPS) with our scene-based non-uniform segmentation. We report conventional QA accuracy on MLVU and VideoMME, as well as streaming metrics on temporal video grounding and dense video captioning on ETBench in Table~\ref{abclip}. The scene-based segmentation yields superior results in both conventional and streaming settings. This aligns with our motivation that scene-based segmentation preserves more structural video information, which facilitates better model learning and more timely responses.

\vspace{2pt}
\noindent\textbf{Special Token Design.}
\begin{table}[]
    \centering
    \small
    \setlength{\tabcolsep}{4pt}
    \begin{tabular}{lccccc}
    \toprule
        Clip & MLVU & V-MME & TVG$_{F1}$ & DVC$_{F1}$ & DVC$_{Sim}$ \\
        \midrule
        Uniform & 59.8 & 55.4 & 34.5 & 33.1 & 18.1 \\
        Scene-based & 61.7 & 57.2 & 36.1 & 33.8 & 18.9 \\
        \bottomrule
    \end{tabular}
    \caption{Ablation study on the clip segmentation. We compare uniform 16-frame clip segmentation and our scene-based segmentation with SigLip.}
    \label{abclip}
\end{table}
\begin{table}[]
    \centering
    \small
    \setlength{\tabcolsep}{4pt}
    \begin{tabular}{cccccc}
    \toprule
        $\langle \text{ANS} \rangle$ & $\langle \text{TODO} \rangle$ & $\langle \text{SILENT} \rangle$ & TVG$_{F1}$ & DVC$_{F1}$ & DVC$_{Sim}$ \\
        \midrule
        \xmark & \xmark & \xmark & 20.1 & 19.7 & 12.3 \\
        \xmark & \xmark & \cmark & 26.3 & 24.9 & 13.1 \\
        \cmark & \xmark & \cmark & 35.2 & 31.0 & 17.2 \\
        \xmark & \cmark & \cmark & 28.7 & 25.6 & 14.5 \\
        \cmark & \cmark & \xmark & 35.5 & 30.2 & 16.8 \\
        \cmark & \cmark & \cmark & 36.1 & 33.8 & 18.9 \\
        \bottomrule
    \end{tabular}
    \caption{Ablation study on the special token designs. We respectively explore the effectiveness of $\langle \text{ANS} \rangle$ and $\langle \text{TODO} \rangle$ in streaming video processor as well as $\langle \text{SILENT} \rangle$ in the final LLM.}
    \label{abtoken}
    \vspace{-10pt}
\end{table}
We also ablate three special tokens in our architecture, i.e., $\langle \text{ANS} \rangle$ and $\langle \text{TODO} \rangle$ in the streaming video processor, $\langle \text{SILENT} \rangle$ token in the final LLM. These three special tokens mainly impact the streaming video interactions, so we report the streaming evaluation metrics on temporal video grounding and dense video captioning in Table~\ref{abtoken}. There are three observations. \emph{First}, the absence of $\langle \text{ANS} \rangle$ token can lead the model to be unaware of the timestamps at which a response has been given. As a result, if relevant clues have appeared in the video contexts, the model tends to produce a response, resulting in a high recall but a low accuracy. \emph{Second}, the $\langle \text{TODO} \rangle$ serves as an indicator that reminds the streaming processor to decide whether to respond. The performance slightly degrades without this special token. \emph{Third}, the $\langle \text{SILENT} \rangle$ token in the final LLM serves as a secondary filter for response decision. If the preceding streaming processor incorrectly identifies a timestamp as requiring a response, the $\langle \text{SILENT} \rangle$ token enables the LLM to rethink whether an answer is needed referring to the historical QA interaction contexts.

\section{Conclusion} 
In this work, we introduced Dispider, a novel framework designed to enable active real-time interaction with video LLMs. By disentangling the key capabilities of perception, decision, and reaction, and adopting an asynchronous processing approach, Dispider overcomes the inherent conflicts that hinder real-time interaction in traditional video LLMs. Our approach ensures that the model can continuously process the video stream while providing timely, contextually accurate, and precise responses to user interactions. The Proactive Streaming Video Processing module optimizes the video analysis by focusing on relevant content, while the Precise Interaction module generates detailed responses asynchronously, allowing for uninterrupted video processing.
We demonstrated the effectiveness of Dispider through extensive experiments on both conventional and streaming video QA benchmarks, where it outperformed existing methods in proactive response capabilities, temporal awareness, and computational efficiency. 
Dispider’s disentangled architecture and ability to handle long-duration video streams make it an ideal solution for real-time interactive applications.

{
    \small
    \bibliographystyle{ieeenat_fullname}
    \bibliography{main}
}


\end{document}